\documentclass[letterpaper, 10 pt, conference]{ieeeconf}  

\IEEEoverridecommandlockouts                              

\usepackage{graphicx}
\usepackage{url}

\title{\LARGE \bf
A localization approach for autonomous underwater vehicles: A ROS-Gazebo framework
}

\author{Frederico C. Vaz$^{1,2}$, David Portugal$^{2}$, Andr\'{e} Ara\'{u}jo$^{2}$, Micael S. Couceiro$^{1,2}$ and Rui P. Rocha$^{1}$
\thanks{This work was supported by the Institute of Systems and Robotics (ISR) – University of Coimbra funded by ``Funda\c{c}\~{a}o para a Ci\^{e}ncia e a Tecnologia`` (FCT) under Grant No. UID/EEA/00048/2013.}
\thanks{$^{1}$F. Vaz, M. S. Couceiro and R. P. Rocha are with Institute of Systems and Robotics, University of Coimbra, 3030-290 Coimbra, Portugal
        {\tt\small \{frederico.vaz,micaelcouceiro,rprocha\}@isr.uc.pt}}%
\thanks{$^{2}$F. C. Vaz, D. Portugal, A. Ara\'{u}jo and M. S. Couceiro are with Ingeniarius, Ltd., 3025-307 Coimbra, Portugal
        {\tt\small \{frederico,davidbsp,andre,micael\}@ingeniarius.pt}}%
}

\begin{document}

\maketitle
\thispagestyle{empty}
\pagestyle{empty}

\begin{abstract}

Autonomous Underwater Vehicles (AUVs) have the ability to operate in harsh underwater environments without endangering human lives in the process. Nevertheless, just like their ground and aerial counterparts, AUVs need to be able to estimate their own position. Yet, unlike ground and aerial robots, estimating the pose of AUVs is very challenging, with only a few high-cost technological solutions available in the market. In this paper, we present the development of a realistic underwater acoustic model, implemented within the Robot Operating System (ROS) and the Gazebo simulator framework, for localization of AUVs using a set of water surface robots, time of flight of underwater propagated acoustic waves, and a multilateration genetic algorithm approach.   

\end{abstract}

\section{INTRODUCTION}

Autonomous Underwater Vehicles (AUVs) have been recommended for scientific exploration and sub-aquatic inspection tasks. These robots need to face harsh environmental conditions of oceans and rivers where visibility is reduced. Many of the mobile robotics technologies are not suitable or do not work correctly under these conditions. Moreover, the high cost of off-shore missions makes it difficult to test hardware and software algorithms on real scenarios. Robotics simulators are a practical and straightforward solution to help developing and testing new ideas and prototypes on a virtual scenario for localization, navigation and control algorithms. A robotic simulator for underwater scenarios should be able to simulate robot motion, sensors and environmental interactions of the AUVs. In addition, the assurance that the developments on a robotic simulator can be realistically transfered to the real robot scenario is an important prerequisite for robotics simulation tools.  

The focus in the present work is on acoustic localization for underwater ROS-enabled robots. There are some high quality AUVs simulators available~\cite{cook2014survey}, but they still do not cover adequately the AUVs localization using acoustic beacons. The localization method proposed depends on the underwater acoustic channel, therefore an important contribution of our work is on modeling acoustic propagation in the water column. We make use of the Unmanned Underwater Vehicle Simulator~\cite{Manhaes_2016}, a set of Gazebo plugins~\cite{koenig2004design} and ROS nodes under development, that are used for supporting and testing the simulation of unnamed underwater vehicles (UUV), i.e AUVs. The UUV Simulator is based on Gazebo, which provides a recognized physical fidelity, a programmatic interface, sensor modeling and adequate documentation.   

There are clear advantages of using the sound channel when compared to the radio channel. Mainly, the radio channel in an underwater environment suffers from large attenuation, resulting in very limited transmission range. The underwater environment is a very complex medium for sound propagation, specially in oceans, because of the heterogeneity and random fluctuations such as the effect of surface seas and the ocean bottom variances. When the acoustic signal propagates in the underwater environment there is a decrease in the signal amplitude due to \textit{geometrical spreading}, the distance traveled by the signal, and the \textit{absorption} due to the chemical properties of the sea or river water. The range of acoustic waves are even more limited in high frequencies due to the \textit{absorption} factor. Additionally, the speed of acoustic waves varies spatially in the sea water, due to variations in temperature and pressure. 

\section{Modeling the underwater acoustic channel}

\begin{figure}[!t]
	\centering
	\includegraphics[width=0.8\columnwidth]{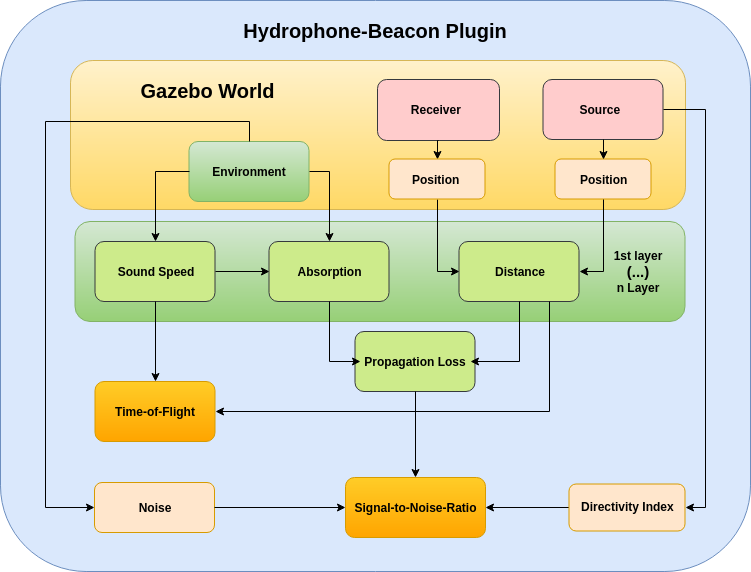}
	\caption{Underwater Acoustic Channel Software Architecture.}
    \label{fig1}
\end{figure}

We modeled the acoustic channel by relating the local values of propagation wave direction (Snell-Descartes law) and velocity. The modeling technique used in our work is known as \textit{geometrical acoustics}. Our method assumes that the water column is divided into several layers that represent different depths with distinct temperature, salinity and pH. 

\begin{figure*}[t] 
	\centering
    \includegraphics[width=0.51\textwidth]{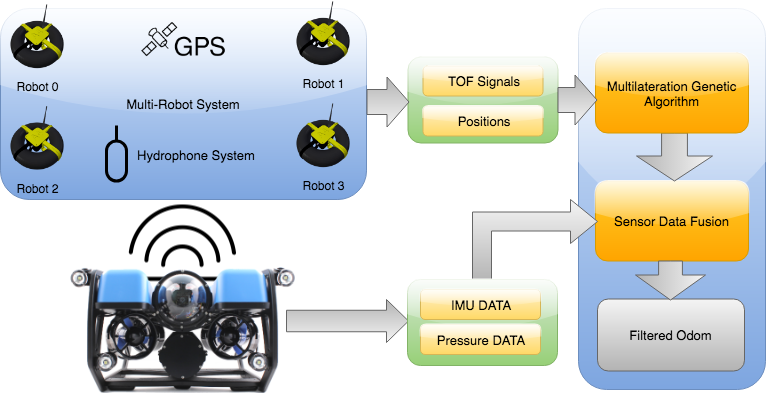} 
	\caption{Localization System Architecture.}
    \label{fig2}
\end{figure*}

\begin{figure}[ht]
	\centering
	\includegraphics[width=0.8\columnwidth]{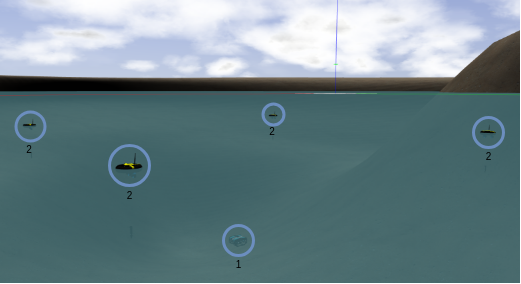}
	\caption{The proposed system in a simulation scenario: The AUV (1) and the 4 water surface robots (2).}
    \label{fig3}
\end{figure}

In the first place, the method gets the source (beacon sensor) and receiver (hydrophone sensor) positions to compute the ground truth distance. Then, the method uses the environment conditions to compute the sound speed and the absorption coefficient for each layer between the receiver and source depth. Thus, the time-of-flight (TOF) between the source and the receiver takes into account the distance and sound speed in each layer. The signal-to-noise-ratio is the result from the environment noise and signal propagation that depends on the absorption coefficient in each layer. The model architecture can be seen in Fig.~\ref{fig1}. 

\section{Description of the system}

\begin{figure}[t]
	\centering
	\includegraphics[width=.8\columnwidth]{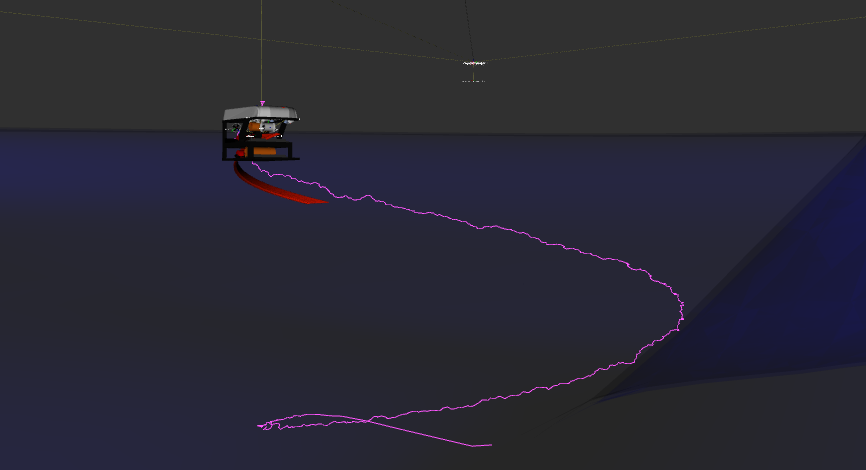}
	\caption{AUV trajectory given by the system.}
    \label{fig4}
\end{figure}

The proposed system for AUV localization comprises the use of four water surface robots as reference points, as shown in Fig.~\ref{fig3}, and receptors of an acoustic signal (through a hydrophone sensor) emitted by a beacon installed on the AUV. With the time of flight (TOF) of the acoustic signal, we are able to estimate the distances between each water surface robot to the AUV. Based on the distance measurements to the AUV and on the water surface robots positions given by the global position system (GPS), we developed a multilateration genetic algorithm method to determine the 3-D position of the AUV. In this way, the localization system will be tracking the position of the underwater robot in earth-centered, earth-fixed (ECEF) to a East North Up (ENU) reference frame. We used data from an inertial sensor (IMU) and a pressure sensor mounted on the AUV to provide global orientation and reduce the localization error of the AUV through sensor data fusion, using an Extended Kalman filter~\cite{MooreStouchKeneralizedEkf2014}. The system architecture and the position given by the system can be seen in Fig.~\ref{fig2} and in Fig.~\ref{fig4}, respectively.

\section{Conclusion and Future work}

In this study, we develop an underwater localization system based on the UUV Simulator~\cite{Manhaes_2016}. Moreover, we have created a new plugin for Gazebo: the Hydrophone-Beacon plugin, which provides a hydrophone sensor and a beacon as actuator. The implementation of the underwater acoustic channel model allows the UUV Simulator to provide an effective modeling and simulation tool for the underwater robotics acoustic localization. This new Localization ROS package enables testing underwater robot systems in a simulation environment before going to real scenarios. Field experiments are foreseen in the close future, in order to validate the system and the models that we use in the simulator plugin.

\bibliography{references}

\end{document}